\documentclass[10pt,journal]{IEEEtran}% if you need to 

\usepackage[utf8]{inputenc} % allow utf-8 input
\usepackage[T1]{fontenc}    % use 8-bit T1 fonts
\usepackage{hyperref}       % hyperlinks
\usepackage{url}            % simple URL typesetting
\usepackage{booktabs}       % professional-quality tables
\usepackage{amsfonts}       % blackboard math symbols
\usepackage{nicefrac}       % compact symbols for 1/2, etc.
\usepackage{microtype}      % microtypography

\usepackage{algpseudocode}
\usepackage{algorithm}
\usepackage{comment}

\algblock{ParFor}{EndParFor}
\algnewcommand\algorithmicparfor{\textbf{parfor}}
\algnewcommand\algorithmicpardo{\textbf{do}}
\algnewcommand\algorithmicendparfor{\textbf{end\ parfor}}
\algrenewtext{ParFor}[1]{\algorithmicparfor\ #1\ \algorithmicpardo}
\algrenewtext{EndParFor}{\algorithmicendparfor}

\usepackage{graphicx}
\usepackage{subfigure}

\usepackage{amsmath}
\usepackage{amssymb}
\usepackage{mathtools}
\DeclarePairedDelimiter{\ceil}{\lceil}{\rceil}

\usepackage{hyperref}
\usepackage{amsfonts}

\usepackage{amsmath,amsfonts,amssymb,amsthm,epsfig,epstopdf,titling,url,array}

\theoremstyle{plain}

\theoremstyle{definition}

\theoremstyle{remark}

\usepackage{tikz}

\begin{document}
\title{Sampling Streaming Data with Parallel Vector Quantization - PVQ \\
% Sampling Streaming Data to Build Scalable High Accuracy Machine Learning Models
}
% \author{Mujahid Sultan}
% \date{Oct 4, 2022}

% \title{Conference title.}
\author{
\authorblockN{Mujahid Sultan}\\
\vspace{0.05in}
\authorblockA{\emph{mujahid.sultan@mlsoft.ai}}
\vspace{0.05in}
}
\maketitle
\IEEEpeerreviewmaketitle

\begin{abstract}
Accumulation of corporate data in the cloud has attracted more enterprise applications to the cloud creating data gravity. As a consequence, network traffic has become more cloud-centric. This increase in cloud-centric traffic poses new challenges in designing learning systems for streaming data due to class imbalance. The number of classes plays a vital role in the accuracy of the classifiers' built from the data streams.

In this paper, we present a vector quantization based sampling method, which substantially reduces the class imbalance in data streams.  
We demonstrate its effectiveness by conducting experiments on network traffic and anomaly dataset with commonly used ML model building methods — Multilayered Perceptron on TensorFlow backend, Support Vector Machines, K-Nearest Neighbour, and Random Forests. We built models using parallel processing, batch processing, and randomly selecting samples. We show that the accuracy of classification models improves when the data streams are pre-processed with our method. We used out of the box hyper-parameters of these classifiers as well as used auto-sklearn\footnote{https://www.automl.org/} for hyper-parameter optimization. 

\end{abstract}


\section{Introduction}

With more and more enterprise data moving to the cloud, it has attracted enterprise applications and traffic to the cloud. This accumulation of data in the cloud and its gravity has resulted in an enormous volume of cloud-centric traffic, which poses new computational challenges in developing machine learning (ML) models from these data streams. We observe that when ML models are built from data streams, they do not perform well as the number of observations per class differs significantly in these data streams.
%In the past decade, several persistent stream processing systems, like Kafka \cite{kreps2011kafka} and Redis \cite{Redis} emerged, as the traditional persistent data management systems could not keep up with the demands of stream processing. In the recent years, due to increasing traffic in the cloud, these persistent stream processing systems, too, could not scale well with the new demands. Therefore, distributed stream processing systems are emerging, e.g., the following Apache projects: Samza \cite{apachesamza}, Spark \cite{apachespark}, Storm \cite{apachestorm}, and Flink \cite{apacheflink} and Google's cloud data flow project \cite{akidau2015dataflow}. 

Class balancing techniques like ROSE~\cite{lunardon2014rose} and SMOTE~\cite{chawla2002smote}, with its variants given in \cite{chawla2003smoteboost}, are generally used to address the class imbalance. In our experiments, we observe that even if the data streams are balanced (i.e., when the number of observations per class is roughly the same) with the methods given in \cite{lunardon2014rose,chawla2002smote,chawla2003smoteboost}. We found that the classification models do not perform well unless some intelligent stream pre-processing or sampling is done.

%This dataset is widely used for building classification models for network traffic and anomaly detection. The practical value of this dataset, 20 years after its publication, is questionable (as the types of attacks have evolved). Yet, the dataset still poses a formidable challenge for ML tools, as the test set  is from a different domain than the train set and has more classes. 

When a dataset is balanced, trivial random sampling may yield good results~\cite{kearns1998efficient}. However, when the dataset is imbalanced, random sampling fails to capture full classes in a dataset~\cite{kearns1998efficient}. 
%A lot of use-cases (ranging from fraud detection in credit card transactions~\cite{chan1998toward}, to cybersecurity threat identification~\cite{cup1999dataset}, to search for elementary particles in Hadron Collider data~\cite{baldi2014searching}) warrant processing of imbalanced datasets with very few instances of the minority classes and several millions of the majority classes.

One can improve ``vanilla'' random sampling by performing random sampling from each class of data independently. Hybrid approaches, such as SMOTE and ROSE down-sample the majority class and up-sample the minority classes. These approaches require knowledge of the labels limiting these to supervised learning. Besides, we find that no substantial modeling accuracy can be achieved by applying these methods to streaming data.

\section{Background and Motivation}\label{problem-statement}
In this publication, we propose a novel method to sample imbalanced streaming datasets for building high accuracy ML models. To demonstrate this on network traffic data streams, we select a labeled intrusion and anomaly detection dataset. There are not enough labeled anomaly datasets publicly available out there. Mainly because organizational Security Information and Event Management Systems (SIEMS) are domain-specific and depend upon many factors, e.g., time, geographical location, and nature of the business to label the anomalies, above all, attack and anomaly labels are proprietary. Therefore, to generalize the problem, we use a publicly available and well researched labeled intrusion detection dataset KDDCUP~\cite{cup1999dataset}. Though there are newer datasets available at \cite{phua2010comprehensive, ring2017flow, sharafaldin2018toward}, these are either smaller or not as widely used by the research community. KDDCUP dataset is a perfect fit for our purposes as it is highly imbalanced; the test set is not from the same domain as the train set, making it very close to real-world problems.

A general rule in ML is --- the more the data, the better learning by ML algorithms. Therefore, we use the full KDDCUP dataset as a single streaming window to create a baseline by first analyzing the entire dataset in a single batch, so that the results of small data streams (\textit{mini batches}) can be compared. We perform a number of ad-hoc experiments. Our testbed is a computer with two Intel Xeon(R) E5-2680v4 processors (allowing execution of 56 hyper-threads in parallel) and 512 GB of memory.

To create a baseline, we use the \textit{full batch} of KDDCUP dataset and trained Naive Bayes~\cite{john1995estimating} model using 10-fold cross-validation using Weka v.3.8.2, and it takes $\approx$30 seconds and another $\approx$100 seconds for testing. The performance of the model, not surprisingly, is not stellar (\textit{precision} of $\approx 0.83$ and \textit{recall} of $\approx 0.70$). We also tried to train support vector machine (SVM) model using Weka on the same dataset, but it could not complete after 15 days, so we have to terminate the training.

We then switch to an automated machine learning tool autosklearn~\cite{NIPS2015_5872}, which automates data pre-processing and parameter optimization for machine learning algorithms. Hyper-parameter optimization and automated machine learning is not a new concept and has been experimenting for several decades. A comprehensive survey is given by~\cite{zoller2019survey}. We set the system memory limit available memory (512 GB), the rest of the default parameters are used and described in Section~\ref{sec:auto-sklearn}.The training process runs for 15 days, on the system described above, before we kill it without any results.

Then we used approach to build models in parallel by sharding the data and training an ensemble of models (one model per data shard)~\cite{drucker1994boosting}.Training of each member of the group is independent of each other hence the ease of parallelization. We implement this approach using the classification algorithms mentioned above. We find that this approach does not work well for the KDDCUP dataset. No individual classifiers can pick all the classes in the dataset, resulting in very low\textit{precision}, details are given in Section~\ref{sec:mlp}.

Then we tried parallel versions of classification models. Though there are ways to perform parallel processing of all the classification algorithms, not all the classification algorithms, have parallel versions available \cite{he2010parallel}. Most of these methods find work-around and either compromise on accuracy or perform some sort of data reduction to design the parallel version of an algorithm, e.g., random forest (RF) is a parallelizable algorithm, but the implementations often compromise on accuracy or performance \cite{chen2017parallel}.
As we discussed above, not every model is parallelizable (even if it is, it may be challenging to do it efficiently, not to mention the cost associated with parallelization). Thus, this approach is not universal.
%SPRINT~\cite{shafer1996sprint}, SLIQ \cite{mehta1996sliq} and ScaleParC \cite{joshi1998scalparc} are few examples of the parallel processing of the classification algorithms, which provide speedup of the processing but lake on the model performance (generalization). 

Finally, we used graph-based parallel processing systems, like TensorFlow~\cite{abadi2016tensorflow}, CNTK~\cite{seide2016cntk}, and Theano \cite{bastien2012theano}, as the backend to speed up the computations on multicore-systems. Not all the classification algorithms can be used with graph-based backends,  and even the ones that can be used, e.g., Neural Networks, are degraded to the extent that these models cannot be generalized. We demonstrate this in detail in Section \ref{results_group2b}.
% Therefore, for the classification algorithms for which the parallel versions are not available or do not generalize well, and single core can not perform the SD, our method can also be used to achieve the speedup and improve accuracy on the commodity hardware. Our method can be used with graph based backend systems like TensorFlow, CNTK and Theno, and produces classification models of very high quality.

%Show the random/sequential read of kdd data and how many classes are present. Then look in to the re balancing the classes and use the sequential/random read and see if we can get some better results. This will lead to the argument that even after class re-balancing the if the distribution does not follow the SQM then the models can not build in parallel and data (in instance space) has to be reduced. 

\section{Proposed Method: Parallel Vector Quantization}
To address the issues discussed in the previous section, we designed a Parallel Vector Quantization (PVQ) method that improves the classification accuracy of the classifiers mentioned above by removing the class imbalance. This method gives much better results when used with graph-based backends like TensorFlow.

\begin{figure}[t]
\centering
    \includegraphics[width = .75\columnwidth]{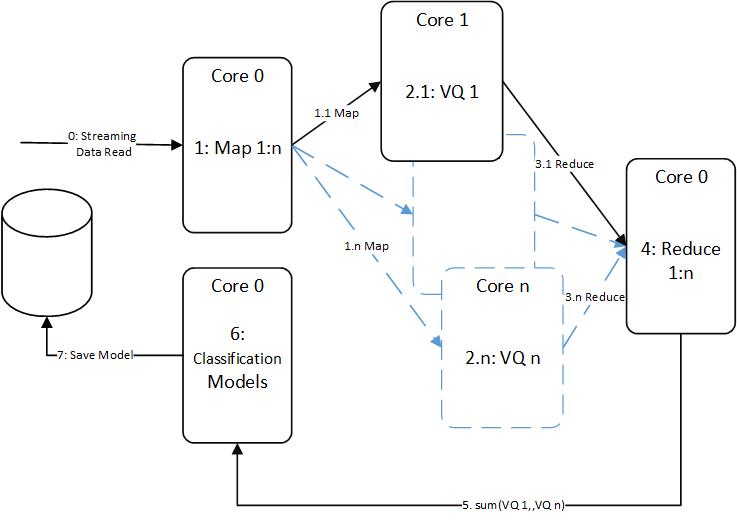}%{pvq01.png}
    \caption{\label{fig:PVQ1}Schematic Diagram of PVQ}
\end{figure}

The schematic diagram of PVQ is shown in Figure~\ref{fig:PVQ1}. The intuition behind this approach comes from scalar quantization. Quantization takes a continuous function, like a $sine$ wave,  samples it at a much coarser scale, and produces a much smaller dataset than the original one. The steps of the scalar quantization are separated by quantization error. In telecommunications, the analog signal is quantized by the coder, and much smaller signal is transmitted across the network, and the decoder then re-constructs it back using the quantization error. Similarly, Vector Quantization (VQ) is a data sampling mechanism in high dimensional spaces that preserves the characteristics of the full dataset. The details of the PVQ method are given in Section~\ref{methods}.

\subsection{PVQ Cloud Architecture}
\begin{figure}[tb]
\centering
    \includegraphics[width=8.cm, keepaspectratio]{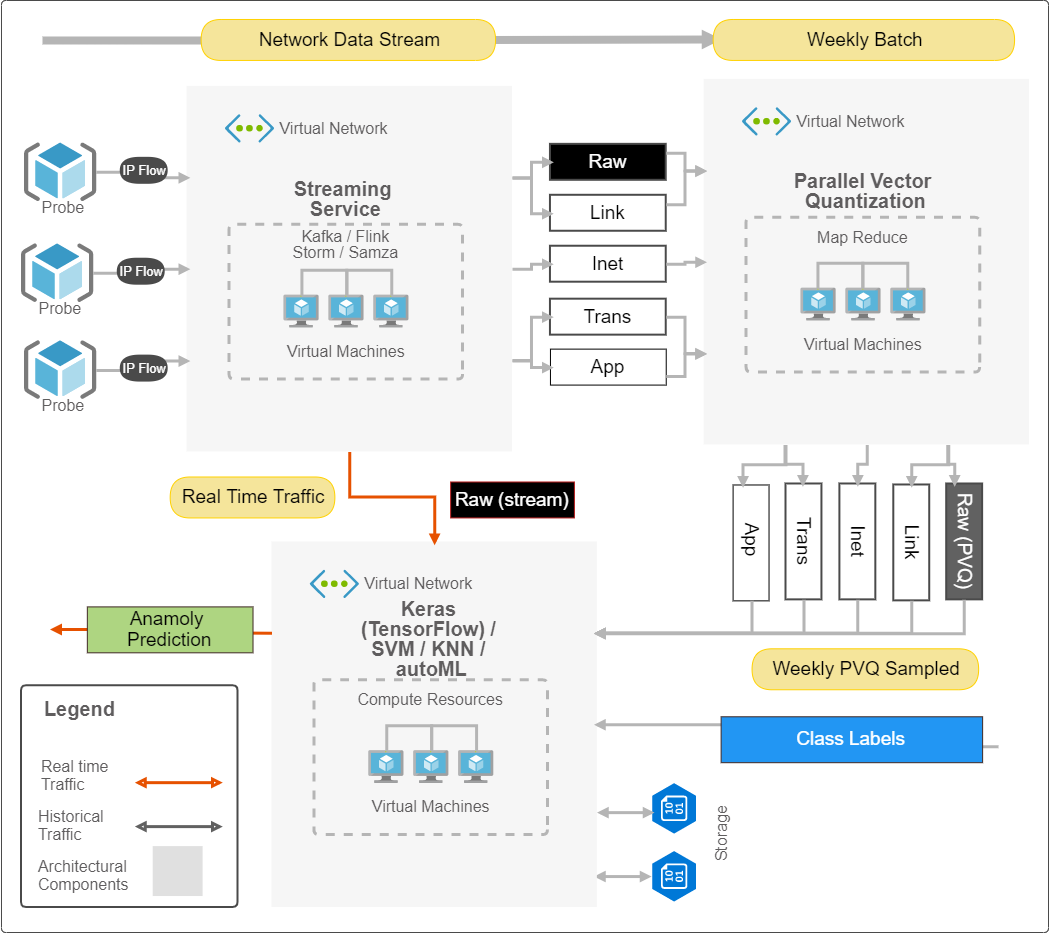}%{pvq01.png}
    \caption{\label{fig:PVQ0} Architecture for real-time classification of streaming data using PVQ in the cloud. PVQ can be used with distributed stream processing systems, as shown in this high-level diagram. Network data streams are captured using Snort or Kafka and, PVQ can be applied to weekly or daily streams. The models build with PVQ sampled data are then used by the classifier to predict anomalies in real-time traffic.}
\end{figure}

\begin{figure}[tb]
\centering
    \includegraphics[width = .7\columnwidth]{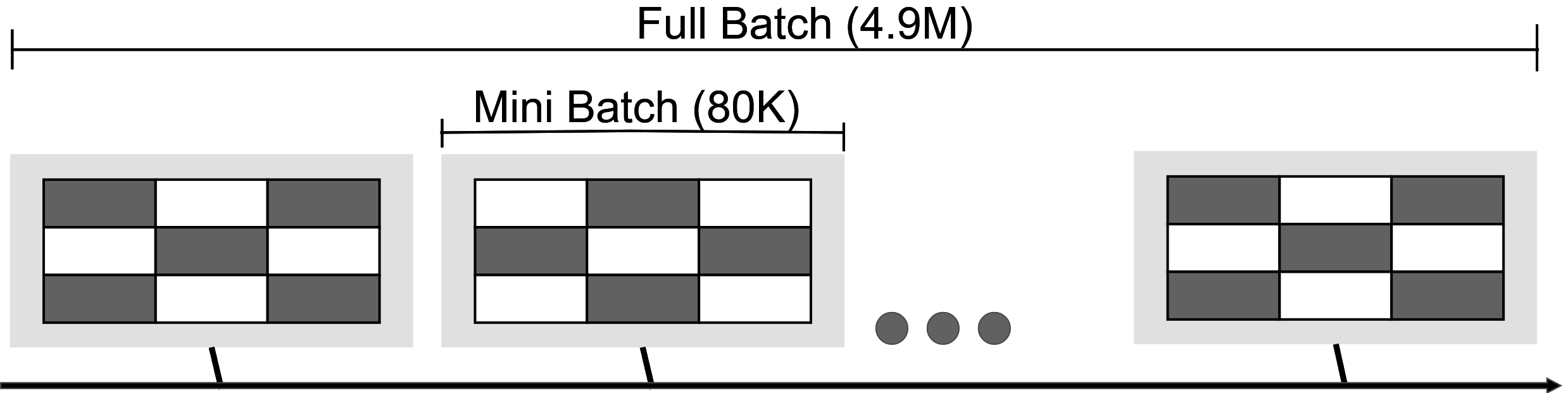}
    \caption{\label{fig:batches} Depiction of full batch and mini-batches for KDDCUP dataset}
\end{figure}

The architecture to use PVQ with distributed stream processing systems in the cloud is shown in Figure~\ref{fig:PVQ0}, and the schematic representation of data streams is shown in Figure~\ref{fig:batches}. \textit{Mini batches} of network traffic data streams are captured by a streaming service Google's cloud data flow \cite{akidau2015dataflow}. Network traffic can be captured by widely used network security monitors (e.g., Zeek~\cite{zeek}) or widely used intrusion prevention systems (e.g., Snort~\cite{snort}). The network traffic data streams are intercepted by the PVQ segment as shown in Figure \ref{fig:PVQ0}. 

The streaming data is sampled by PVQ and passed to ML model building layer. To demonstrate our method, we stream \textit{mini-batches} of 80k network traffic packets and a \textit{full batch} of 4.9M packets, as shown in Figure \ref{fig:batches}. 

\subsection{Our Contributions}

The purpose of our work is not to evaluate how different ML algorithms perform on streaming datasets (for which the reader is directed to \cite{anton2018evaluation}) nor to build a new intrusion detection system, which is a well-research area. The contributions of our work are summarized as follows.
\begin{enumerate}
    \item We show that PVQ removes class imbalance in the streaming data even with a small window;
    \item Most importantly, we show that when graph-based parallel processing backends like TensorFlow are used with classification algorithms, the classifiers cannot capture full classes present in a streaming dataset, even if a very large sampling window is used. Whereas, when data is PVQ sampled the classification algorithms (with TensorFlow backend) captured almost all classes present in a dataset;
    \item We describe a parallel vector quantization (PVQ) sampling method, which improves the accuracy and performance of the classification algorithms for cloud-centric streaming data.
\end{enumerate}

The rest of the paper is structured as follows. In Section~\ref{literature_review} we discuss the related work. In Section~\ref{methods} we describe the algorithm in detail and its complexity analysis to compare with similar methods to run in real-time stream processing. The experimental setup is given in Section~\ref{experimentalsetup}. Results for experiments are discussed in Sections~\ref{results} and \ref{discussion}. 

\section{Literature Review}\label{literature_review}

Machine learning model building from streaming data is a well-researched area \cite{gama2010knowledge, gaber2005mining, pratama2014pclass}, a survey of clustering methods from data streams can be found at \cite{silva2013data}. A survey on addressing the class imbalance in big data is given by \cite{leevy2018survey}. Incremental learning and concept drift has been discussed by \cite{mehta2017concept, pratama2016incremental,sayed2012learning}, and a survey on \textit{concept drift} adoption can be found here \cite{gama2014survey}.

Spark ML engine is used by~\cite{bifet2015streamdm}.   Long short term memory neural networks have been used to detect patterns from streaming data \cite{malhotra2015long}, but they do not discuss the class imbalance. A survey on deep learning with class imbalance is given by \cite{johnson2019survey}. Maximum likelihood estimation has been used in anomaly detection in streaming data by \cite{akouemo2016probabilistic}.
A theoretical overview of the methods to build ML models from large datasets can be found in \cite{al2015efficient, xing2015petuum}. Theoretical details of the state-of-the-art techniques to learn from imbalanced data are given in~\cite{he2008learning}.

Most conventional network sniffers or network security monitors like Zeek~\cite{zeek} and intrusion prevention systems like Snort~\cite{snort} are used in combination with distributed streaming engines (e.g., Spark \cite{apachespark} and Flink \cite{apacheflink}) to gather the network statistics but none of these address class imbalance in streaming data. Performance comparison of distributed network streaming engines is given by \cite{chintapalli2016benchmarking}.

None of these studies explore sampling or addressing the class imbalance in streaming data using VQ. Vector quantization is widely used as a mechanism of dimensionality reduction in high dimensional and multivariate domains \cite{gersho2012vector}. Background on VQ can be found in \cite{gray1998quantization}. VQ methods like Dirichlet tessellations \cite{dirichlet1850reduction} and Voronoi tessellations \cite{voronoi1908nouvelles} date back to the ninetieth century. And in the modern times introduced by Lloyd's \cite{lloyd1982least}  (in a scalar form) and Forgy \cite{forgy1965cluster} in vector form. VQ combined with SMOTE has been used by \cite{nakamura2013lvq} as supervised learning in the biomedical domain and lacks a discussion on unsupervised and streaming datasets.VQ in streaming video data was used by \cite{khoshrou2015learning} but did not consider class imbalance.

%The prime task of some SD methods, like Principle Component Analysis (PCA) \cite{jolliffe2002principal, roweis2000nonlinear}, Support Vector Machine (SVM)~\cite{boser1992training, cristianini2000introduction, tayal2014primal}, and t-SNE~\cite{maaten2008visualizing} is to reduce dimensionality while preserving the global structure in the data. t-SNE preserves local data structures, whereas PCA and SVM  take a projection of the data to a suitable or favorable dimension \cite{roweis2000nonlinear}. The resulting dimensionality reduction may help in visualization of the data, but if the reconstruction of the original data is required then these models are not well suited for the task~\cite{roweis1998algorithms}. VQ on the other hand, is a  better mechanism of dimensionality reduction in high dimensional and multivariate domains \cite{gersho2012vector}. Background on VQ can be found in \cite{gray1998quantization}. VQ methods like Dirichlet tessellations \cite{dirichlet1850reduction} and Voronoi tessellations \cite{voronoi1908nouvelles} date back to ninetieth century. And in the modern times introduced by Lloyd's \cite{lloyd1982least}  (in scalar form) and Forgy \cite{forgy1965cluster} in vector form. 

\section{Parallel Vector Quantization (PVQ)}\label{methods}

\subsection{Algorithm}\label{PVQalgo}

Our algorithm is summarized in Algorithm~\ref{alg:pvq}, the details are as follows.
Let input $X=\{x_1, \ldots , x_n\}$ be a dataset composed of $n$ instances; and $i$-th instance $x_i \in \mathbb{R}^d, i = 1, \ldots, n$. 

To perform parallel processing we randomly partition $X$ into $L$ subsets $S=\{s_1, \ldots, s_L\}$. We then independently select representative samples (using VQ) of points from each of the $L$ subsets. We deem such a representative sample for subset $s_k$ as $\pi_k$.

This is an embarrassingly parallel problem, as the selection of $\pi_k$ within each of the $L$ subsets is independent of the remaining subsets. Thus, the selection can be performed in parallel independently for each of the $L$ subsets. Once we obtain $\pi_k$ from each of the $L$ susbsets we create a union of all $\pi_k$, producing the final representative sample $\Pi = \bigcup\limits_{k=1}^{L}{\pi_k}$.   The resulting dataset $\Pi$, which is much smaller than $X$, is passed to classification systems for modeling. 

In the MapReduce programming model, the selection of $\pi_k$ in each of the $L$ subsets is done in the Mapping phase, followed by the union of $\pi_k$  for all $L$ in the Reduction phase. Details of the phases are as follows:

\subsubsection{Map}
For $k$-th subset of data $s_k$, a VQ algorithm finds a set of codebook vectors or centroids $C_k$ with $m_k$ elements: $C_k =\{c_1, \ldots , c_{m_k}\}$, with $j$-th instance $c_j \in \mathbb{R}^d, j = 1, \ldots, m_k$. Typically,  $ |s_k| \gg m_k $, where  $|s_k|$ is the number of elements in $s_k$.

Each observation in $s_k$ belongs to a particular centroid. The mapping (in the mathematical sense of the term) of observations in $s_k$ to codebook vectors in $C$ is done by computing Voroni cells. That is, a point $x_i$ belongs to $c_j$ if the Euclidean distance between $x_i$ and $c_j$ is shorter than the distance between $x_i$ and the remaining $m-1$ centroids in $C_k$. Once the mapping, deemed $\mathcal{M}_k$, is established, for every centroid that has at least one observation in $s_k$ mapped to it, we retain an instance that is the closest to a given centroid. The set of the points that are the closest to the centroids, becomes our representative sample of $s_k$. We deem this sample set $\pi_k$.

Various VQ algorithms can be used. We use Self-Organising Maps (SOM) \cite{vesanto1999self}, as it is fast and scalable algorithm that can deal with high-dimensional data~\cite{kohonen2013essentials}. To execute SOM, we need to define topology of SOM which is done using standard heuristic given by~\cite{vesanto2000neural}. The heuristic sets  $m_k$ proportional to $|s_k|$: 
\begin{equation}\label{eq:mk}
m_k = \ceil*{5 \sqrt{|s_k|}}. 
\end{equation}
Typically, $|s_k| \approx n / L$. 

\subsubsection{Reduce}
Append all $\pi_k$ (computed in parallel during the Mapping phase) to produce the resulting set $\Pi$. 

\subsubsection{Analysis of $\Pi$ }
Given that we partition $X$ into $L$ subsets with approximately the same number of elements, the number of elements in $k$-th subset $s_k$ is $\approx n/L$. The number of elements in ${\pi}_{k}$ will be at most $m_k$. Given Eq.~\ref{eq:mk}, the upper boundary of the elements in $\Pi$ is

\begin{equation}\label{eq:pisize}
|\Pi| = \sum_{k=1}^{L}{m_k} = \sum_{k=1}^{L}{\ceil*{5\sqrt{|s_k|}}} \approx \sum_{l=1}^{L}{5\sqrt{n/L}} = 5\sqrt{nL}.
\end{equation}
Eq.~\ref{eq:pisize} helps us estimate the number of samples returned by the PVQ. We can control the size of $\Pi$ by changing the number of subsets $L$.

\begin{algorithm}
\caption{Parallel VQ (PVQ)} % give the algorithm a caption
\label{alg:pvq} % and a label for \ref{} commands later in the document
\begin{algorithmic}[1]
    \Procedure{pvq}{$X,L$} %\Comment{See Section~\ref{PVQalgo} for definitions of $X$ and $L$}
        \State{$S \gets$ split the dataset $X$ into $L$ subsets}
        \State{$\Pi \gets$ an empty set}
        \ParFor{$s_k$ in $S$}  \Comment{ Mapping phase, which is done on the worker nodes;  $k=1,\ldots,L$.}
            \State{$\pi_k \gets$ an empty set}
            \State{$m_k \gets \ceil*{5 \sqrt{|s_k|}}$ } \Comment{The scalar $m_k$ is computed as per~\cite{vesanto2000neural}}
            \State{$T_k \gets$ define SOM topology using~\cite{vesanto2000neural} based on $m_k$}
            \State {$C_k, \mathcal{M}_k  \gets$ Compute VQ of $s_k$ using SOM based on $T_k$}
            \For{$c_j$ in $C_k$} \Comment{ $j =1, \ldots, m_k$ }
             \State{$r_k \gets$ a subset of observations belonging to $c_j$ \Comment{obtained using $\mathcal{M}_k$}}
                \If{$r_k$ is not empty}
                    \State{$o \gets$ an observation in $r_k$ that is the closest to $c_j$ (based on the Euclidean distance) }
                    \State{Add $o$ to $\pi_k$} 
                \EndIf
                \State{Add $\pi_k$ to $\Pi$} \Comment{Reduction phase: in practice, this step will be done by the master node}
            \EndFor
            
        \EndParFor
        \State \textbf{return} $\Pi$
    \EndProcedure
\end{algorithmic}
\vspace{-1pt}
\end{algorithm}

The Voronoi region $V_i$ of the codebook vector $c_i$ is the set of vectors in $\mathbb{R}^d$ for which $c_i$ is the nearest vector: $V_i = \{z \in \mathbb{R}^d |i =$ arg min ${||z - c_j||}^2\}$. 
The definition of the Voronoi set $\pi_i$ of the codebook vectors $c_i$ is straightforward. It is the subset of $X$ for which the codebook vector $c_i$ is the nearest vector: $\pi_i = \{x \in X|i = $ arg min ${||x-c_j||}^2$ that is, the set of vectors belonging to $V_i$. %A partition on $\mathbb{R}^d$ induced by all Voronoi regions is called Voronoi tessellation or Dirichlet tessellation.

Data streams are read from distributed streaming systems, and VQ is performed in parallel using MapReduce. The task of VQ is to find $\pi_i$ for each subset of the data. Let us assume that we have $w$ workers and that the data is partitioned into equal size shards between the workers. Then each worker in the system reduces $n/w$ instances to $m$ codebook elements. The codebooks from all $w$ workers are appended together, resulting in the sampled dataset $X'={\pi_1, \ldots , \pi_w}$ of the size $\leq mw$ (in practice, some of the codebook records can be empty, thus $mw$ represents an upper boundary of the size of the reduced dataset). The resulting dataset $X'$, which is much smaller in size, is passed to classification systems for modeling. 

This process is formally depicted in Algorithm~\ref{alg:pvq}. To perform VQ, we use a batch version of SOM Toolbox \cite{vesanto2000neural}. Note that our approach does not require labels; thus, it can be be used in supervised and unsupervised learning pipelines.

Each worker will take In each iteration $iter$ each worker produces a set $C_{iter}$ where $iter =$ number of workers which participated in the computations. This reduced the dataset $X$ into a much smaller dataset $X'={x'_1}, \ldots , x'_{iter}$. as given in Algorithm \ref{alg:pvq}. 

To perform VQ, we used a batch version of SOM Toolbox \cite{vesanto2000neural}. The resulting dataset $X'$ which, is much smaller in size, is used for classification model building, and we show that this set substantially improves the class imbalance and classification accuracy of the models built from this reduced dataset.  

Results of different datasets show that this scheme is very efficient, reduces complexity, and produces very good predictive models as if the models were built on the full dataset.

\subsection{Complexity Analysis}\label{sec:complexity}
The theoretical benefit of using PVQ can be sketched as follows. Let us assume that Algorithm~\ref{alg:pvq} will be executed on $w$ workers. Without the loss of generality, and to simplify computations, let us assume that the number of workers is equal to the number of data subsets, i.e., $w=L$. The dataset $X$ will be distributed among the workers in roughly equal shards. Then, each worker has to process $\approx n/L$ observations.

VQ drives PVQ worst-case complexity at each worker. Worst-case VQ complexity on a full dataset is $O(dn^2)$~\cite{gersho2012vector}. However, since each of the workers is processing only $n/L$ instances, the complexity is reduced to $O\left( d L (n/L)^2 \right) = O\left( d n^2 / L \right)$.

SVM's worst-case computational complexity is $O(dn^3)$~\cite{bishop2012pattern}. The complexity of processing the reduced dataset $\Pi$ is $O(d |\Pi|^3) \ge  O(d (nL)^{3/2}) $. Besides, we need to add the time needed to reduce the original dataset: $O\left( d n^2 / L \right)$. This will yield $O\left(d (nL)^{3/2} + d n^2 / L \right)$. Given that $n \gg m > L$, we may assume that $d n^2 / L$ will dominate the $d (nL)^{3/2} $ term (i.e., $n^2 \gg d (nL)^{3/2} $). Thus, the complexity will be mainly driven by the PVQ-related term $O\left( d n^2 / L \right)$.

A similar analysis can be performed for k-nearest neighbors (kNN) with the worst-case complexity of $O(dn)$~\cite{aha1991instance} and random forest (RF) with the worst-case complexity of $O\left(d n^2 \log(n)\right)$~\cite{breiman2001random}, and multilayer perceptron (MLP) with the worst-case complexity of $O\left(2^{d-1}n\right)$~\cite{rumelhart1985learning}.

\section{Experimental Setup}\label{experimentalsetup}

\subsection{Dataset Description}\label{sec:dat_description}
KDDCUP is a classic dataset widely used for building classification models for network traffic and anomaly detection. This dataset is based on an intrusion detection system evaluation program \cite{lippmann2000evaluating} and collected by \cite{stolfo2000cost}. It contains the data for seven weeks of network traffic. 

While the types of attacks in the dataset may be stale, the dataset's structure is very relevant for today's problems. The dataset contains a large number of observations and classes; some of the classes present in the train set are missing in the test set and vice versa. To quantify, the train set contains 4.9M instances, with 41 features labeled as normal class or one of the 22 attack classes. The test set contains 311K instances and 38 classes (one normal and 37 attack types). Out of these 38 classes, the training dataset has only 23 classes; the remaining 15 classes are not present in the train set. The dataset is highly imbalanced:  the most common class ``smurf'' has 2,807,886 observations, while the least common class ``spy'' has only 2 instances. In these experiments, we want to demonstrate that PVQ pre-processing takes care of the class imbalance by picking maximum number of classes. This is then verified by the higher performance statistics of the classifiers.
\subsection{Data Preparation}
\subsubsection{Transformation of features} We add a dummy feature to the training and test sets to improve the generalization of models (a common practice in machine learning~\cite{bishop2006pattern, mackay2003information}). We also remove three non-numeric features from the dataset for the simplicity of calculations.
\subsubsection{Data preparation for the complete train set} \label{sec:full_prep}
To mimic  larger sampling window, use the full train set.
%We first experiment with a complete train set of observations. 
To combat the class imbalance in the full train set, we use SMOTE algorithm to  (a) up-sample the ``minority'' class giving 64M instances, (b) down-sample the ``not-majority'' class leading to 2M instances, and (c) up-sample the minority class and down-sample the majority class to a instances in ''satan'' class giving us 365K instances.

\subsubsection{Data preparation for a subset of the training data}

To mimic smaller sampling window, we employ two approaches. In the first one, we randomly sample 80K observations from the original unbalanced 4.9M training observations. In the second one, we use PVQ to ingest the full train set and return approximately 80K representative observations that will be passed to models to do the training.

\subsubsection{Aggregation of labels}

The 23 types of labels present in the train set can be grouped into five categories as per~\cite{sahu2014detail}. Often, in the literature, the authors explore the performance of their predictive models by examining the ability of a model to distinguish five labels (i.e., one `normal' label and four `aggregate attack' labels) rather than 23 original labels (i.e., one `normal' label and 22 `original attack' labels). We try to reduce the labels' space to see if it improves the performance of the models.

\subsection{Models and Frameworks under study} \label{sec:models}

The resulting train sets are used to train the following models: MLP, RF , SVM,  and kNN, albeit we are not comparing how these models perform on streaming data for which the reader is referred to \cite{anton2018evaluation}.

\subsubsection{Keras-based Multilayer perceptron (MLP)}\label{sec:mlp}

We constructed an MLP with $\tanh$ activation as the input layer, two hidden layers, and $softmax$ activation at the output layer. For optimization of the gradient, we used Adam optimizer~\cite{kingma2014adam}. The model is built with Keras v 2.3.0 \cite{chollet2015keras} on TensorFlow backend v 2.1. Keras simplifies the creation of neural networks and pre-configures the majority of hyper-parameters with default values (that work relatively well out of the box). While there exists an infinite amount of MLP architectures, this one mimics a frequent one that a practitioner may start with when exploring a problem.

%Design choices for setting the architecture of TensorFlow are as follows: For multi-class classification problems the loss function of choice is  categorical cross-entropy and we use ADAM~\cite{kingma2014adam} learning rate optimizer. To avoid over fitting we use two dense convolutional layers, in the input layer we use \textit{Tanh} activation function as it is bound between ranges (-1,1) so our choice for the multi-class classification. And in the output layer we use \textit(sigmoid) activation function to give us the probability distribution of all classes to be classified.

\subsubsection{auto-sklearn based RF, SVM, and kNN}\label{sec:auto-sklearn} 

For all three models (RF, SVM, or kNN) we auto-tuned their hyper-parameters using AutoML paradigm~\cite{NIPS2015_5872}, which is implemented in the  auto-sklearn package v 0.6.0.

\subsubsection{Scikit-learn-based RF, SVM, and kNN} \label{sec:scikit} 

We also trained each of the three models (RF, SVM, and kNN) individually using their implementations in scikit-learn v.0.21.0 (with the default values of hyper-parameters set by the package).

\subsection{Experiments}
With the `building blocks' discussed above, we set up two groups of experiments. . The first group studies the models' performance on the full dataset and acts as a general baseline. The second group trains the models on subsets of data, comparing our PVQ approach against random sampling. The details are as follows.

\subsubsection{Group 1}\label{group1}
The first group of experiments is conducted on the full dataset, mimicking a larger sampling window and to set a baseline. In this group, we study the performance of the popular implementations of machine learning models using platforms that are popular among practitioners that would like to obtain the results with a minimal amount of customization. Thus, we leverage the auto-sklearn package (as per Section~\ref{sec:auto-sklearn}) to train RF, SVM, and kNN models. We also train Keras-based implementation of the MLP (as per Section~\ref{sec:mlp}). 

%As a baseline, we train the model on the complete dataset of data, which we prepare as per Section~\ref{sec:full_prep}. 

\subsubsection{Group 2}\label{group2}

The second group of experiments is designed to mimic smaller sampling winds of the data streams. In this group, we evaluate the performance of RF, SVM, kNN and Keras based MLP for \textit{Random sampling} vs \textit{PVQ sampling} of 80k data streams. We further divided these experiments to 

 \begin{itemize}
        \item [a)] using auto-sklearn (to auto-optimize classifier hyper-parameters) (as per Section~\ref{sec:auto-sklearn})
        \item [b)] using vanilla hyper-meters of classifiers (as per Section~\ref{sec:mlp} and Section~\ref{sec:scikit})  
    \end{itemize}
%the behaviour of practitioners who may find the performance of the automatic tools from Group 1 inadequate (from the perspective of timing needed to train the model as we will see later in Section~\ref{xxx}). In this case, a typical solution is to reduce the input size of the data by downsampling. This should lead to a reduction in the training time. But will it lead to the degradation of the predictive power of the models? 

%To test this, we retain  Keras-based implementation of the MLP (as per Section~\ref{sec:mlp}), but replace the auto-sklearn based implementation of RF, SVM, and kNN with the scikit-learn-based ones (as per Section~\ref{sec:scikit}), to speed up the training time. While in this case, practitioners will have to train RF, SVM, and kNN individually (unlike in the auto-sklearn case), the overhead associated with manual labour is low, as we are using default hyper-parameter values.

\subsection{Formulation of Performance Metrics for Multi-Class Classification} \label{performance_metrics}

KDDCUP dataset is a multi-class classification problem. Tt is quite challenging to evaluate classification models for this dataset as the training and test sets have different number of classes. In the following, we state the multi-class problem and formulation of performance metrics and their desired values.

Each instance of the multi-class dataset $x_i$; $X = \{x_1, \ldots ,x_n\}$ has an given class label $\lambda(x_i)$; $\Lambda = \{\lambda_1, \ldots , \lambda_k\}$. Each classifier predicts $\psi_i$ of the each $x_i$ giving the prediction set $\Psi = \{\psi_1, \ldots , \psi_k \}$. The $k \times k$ matrix containing the number of actual and predicted classes as its cells, called the confusion matrix, given by $M_{i'j'}$, where $i'=\Psi$ and where $j'=\Lambda$.

All the classifiers that we used produce discrete classification. Therefore, the overall performance of a classifier $\alpha$ is given by the sum of instances on the diagonal of matrix $M$ divided by the number of all instances, called \textit{accuracy}, can be written as: 

\begin{equation}\label{eq:accu}
accu(\alpha) = 1/n \sum_{i'=1}^k M_{i'i'}.   
\end{equation}

The \textit{precision} and \textit{recall} of a classifier is calculated as:

\begin{equation}\label{eq:precision}
precision(\alpha) = 1/n \sum_{i=1}^n \frac{|\lambda (x_i)\cap \psi(x_i)|}{|\psi(x_i)|}    
\end{equation}
and 
\begin{equation}\label{eq:recall}
recall(\alpha) = 1/n \sum_{i=1}^n \frac{|\lambda (x_i)\cap \psi(x_i)|}{|\lambda (x_i)|}.
\end{equation}

% Point to note in Equations (\ref{eq:precision}) and (\ref{eq:recall}) is the denominator. The denominator in \textit{recall} is class label $\lambda(x_i)$, making \textit{recall} sensitive to the total number of class labels in a dataset. Whereas, \textit{precision} is not sensitive to the class labels. High \textit{accuracy} with high \textit{precision} and low \textit{recall} will be the desired results in case of high-class imbalance and extra classes in the test set. 

Area under curve and receiver operator curves are also widely used metrics for assessing classifier \textit{accuracy} but become too cumbersome for multi-model comparison. Balanced Accuracy (BA), on the other hand, combines \textit{accuracy} and \textit{recall} into one metric and gives a good visual comparison of multi-model comparison. Therefore, we compare the models using \textit{accuracy}, \textit{precision}, \textit{recall}, and \textit{Balanced Accuracy}, as these measures provide a very good visual representation of the overall performance and generalization capability of a model. 

For multi-class and imbalanced datasets, generalized Matthews Correlation Coefficient (MCC)~\cite{gorodkin2004comparing} score also gives unbiased results (see \cite{powers2011evaluation, boughorbel2017optimal}) as it involves values of all quadrants of the confusion matrix. 
Therefore, we also use this metric %\cite{boughorbel2017optimal} 
%and given in Equation~\ref{eq:mcc} 
to cross-examine \textit{accuracy}, \textit{precision}, and \textit{recall} combination.

\subsection{Evaluation criteria}

To evaluate each experiment, we compare how many classes were detected in that particular set of data. Then we look at the classification performance and see how the class imbalance has effected an experiment. 

% To evaluate performance of the a classification model, we compute accuracy, precision, recall, and the generalized Matthews Correlation Coefficient (MCC)~\cite{gorodkin2004comparing}. The latter is often used to yield unbiased performance results (see \cite{powers2011evaluation, boughorbel2017optimal}) for highly imbalanced datasets (similar to the one we have under study). 

Each experiment is repeated 50 times, and the distributions or mean, and variance of the performance measures are reported. The only exception is the Group 1 (\ref{sec:mlp}) keras-based implementation of the MLP. For this group, we limit the number of repetitions to five, as a single repetition takes approximately one week, and the variance between runs is low.

\section{Results}\label{results}

\subsection{Results: Group 1}\label{results_group1}
The auto-sklearn was non-scalable, as it could not optimize hyper-parameters of RF, SVM, and kNN in 15 days (after which we terminated the process). 

Keras-based implementation of the MLP, was though trained in $~12$ hours on full dataset, the results are not good for any classification modeling. MLP's goodness of fit is reported in Figure~\ref{fig:Box_plots_comparison}. The training on the original  and the up-sampled datasets yielded relatively high accuracy and recall, $\approx 0.9x$ on average. Balancing the data in all classes (via up- and down-sampling) let to the reduction of accuracy and recall to $\approx 0.8x$ on average. The down-sampling scheme yielded the worst results with the average accuracy and recall of $\approx 0.4x$. Precision for all of these data preparation approaches is low, with the averages ranging between $\approx 0.1x$ and $\approx 0.1x$. The reason for low precision is explained below in Section~\ref{auto-sklearn_low}.

Thus, we can conclude that for the full dataset, we would fail training classic models using an AutoML framework on tbis moderate sized dataset. The Keras-based neural network solution would finish training; however, the goodness-of-fit of the model would be inadequate due to low precision for all class balancing schemes. 
%We will further discuss the root cause of low precision in Section~\ref{desired_metrics}.

\subsection{Results: Group 2(a)} \label{results_group2a}
\begin{figure}[tb]
\centering
\begin{tabular}{l}
     (a)\includegraphics[width = .65\columnwidth]{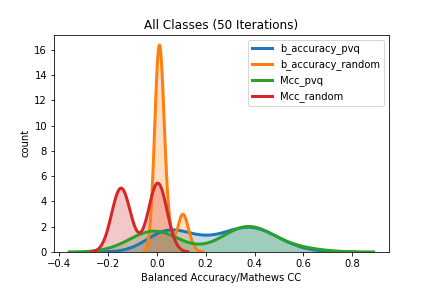} \\
      (b)\includegraphics[width = .65\columnwidth]{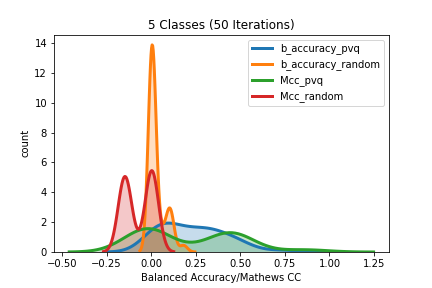}
        \end{tabular}
    \caption{\label{fig:hist_b_accu_5_classes} Balanced Accuracy and MCC for 50 random runs of auto-sklearn. Comparison between PVQ sampled (80k) vs Randomly sampled (80k) (from train set) tested on 10k (from test set) - (a) All 23 Classes in train set (b) Consolidated 5 Classes as given by \cite{sahu2014detail} . Note the b\_accuracy\_pvq and Mcc\_pvq is better in both plots as compared to b\_accuracy\_random and Mcc\_random}
\end{figure}

We sample 80K data points with PVQ and built models for 50 random iterations. Note that we use out of the box auto-sklearn and only specified memory = $512$ MB. We compare the PVQ sampled dataset classification results with the results of equal-sized randomly selected streams. Both models are tested on 10k test set randomly selected from the full test set (which is 311k in total). Balanced Accuracy (BA) and MCC for each of 50 runs is shown in Figure~\ref{fig:hist_b_accu_5_classes}(a). For simplicity of the plot, we select BA over \textit{accuracy} and \textit{recall} as the former takes into account the \textit{recall} of the model. We can see, though PVQ sampled set performed better than the randomly selected subsets (both the BA and MCC are higher - on the horizontal axes), the overall performance of both models is quite low. The reason might be the high number of attack types in the train set (23 in total). 

We performed the same experiment on reduced attack types, as suggested by many studies \cite{tavallaee2009detailed}. Network attacks fall under five main categories. In the literature, numerous studies that build models on these categories. We transformed the 23 class problem into 5 class problem as described by \cite{sahu2014detail}. %and shown in the Table~\ref{tab:5classes}. 
The same classifiers are trained as before using auto-sklearn, and the BA for 5 classes is given in Figure~\ref{fig:hist_b_accu_5_classes}(b). We can see the performance went up as compared to Figure~\ref{fig:hist_b_accu_5_classes}(a), and PVQ is winner, but the \textit{accuracy} is very low, making this method of not much use as well.

\begin{figure}[t]
  \centering
  \begin{tabular}{c}
    \includegraphics[width = .7\columnwidth]{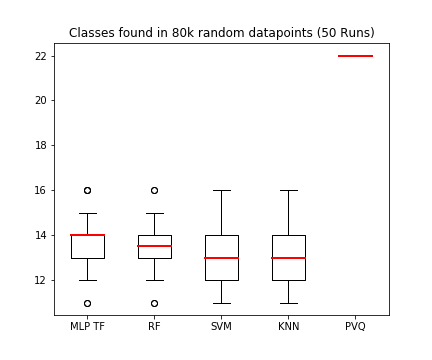}\\
%includegraphics[width=80mm]{Hist-classes.png}\\
   \end{tabular}
    \caption{\label{fig:random_classes} Box plot of the number of train set classes (unique class labels) present in random sampling of 80k from train set for each of classifiers (MLP, RF, SVm and KNN) the left four plots; and the right-hand-side plot: number of train set classes (unique class labels) present in PVQ sampled train set of approximately 79K. (for 50 random iterations)}
\end{figure}

\textbf{Reasons for low \textit{accuracy} by auto-sklearn:}\label{auto-sklearn_low}  
The reasons for low \textit{accuracy}by auto-sklearn  are twofold. First, the test set has many more classes than the train set. Second, auto-sklearn uses sequential model-based algorithm configuration SMAC \cite{smac}, in which hyper parameters optimization process is terminated if it does not converge after a certain time (giving zero \textit{accuracy}). The reason PVQ sampled dataset performed better than the randomly selected is that PVQ at average picked 22 out of 23 classes present in the train set. In contrast,  the random subset of 80k picked a maximum of 16 classes as shown in the Figure~\ref{fig:random_classes}. 
%The results of timing experiments in Group 2 are given in Figure~\ref{fig:timing_grp2}. The training time decreased dramatically, in comparison with Group 1 training time: from hours and days to one minute or less. The process of sampling the data by PVQ, followed by models' training, is slower than the combination of random sampling and training. This is expected: PVQ sampling, by construction, is a more laborious process than random sampling. However, the results are comparable  magnitude-wise, which is achieved by the parallelization of the PVQ.

% Let us now analyze the goodness-of-fit.

% Group 1 vs Group 2: to be done
\subsection{Results: Group 2(b)} \label{results_group2b}

In this group, we used scikit\-learn (sklearn)~\cite{pedregosa2011scikit} machine learning library for calculating performance metrics \textit{accuracy}, \textit{precision}, and \textit{recall}. Whereas we had few design choices to make for hyper-parameter selection of these performance metrics. 

\textit{Accuracy} is not sensitive to the number of test set class labels. Consider the denominator in Eq.~(\ref{eq:precision}) and Eq.~(\ref{eq:recall}). The denominator in the \textit{recall} is class label $\lambda(x_i)$, making \textit{recall} sensitive to the total number of class labels in a dataset. Whereas \textit{precision} is not sensitive to the class labels. High \textit{accuracy} with high \textit{precision} and low \textit{recall} will be the desired results in case of high-class imbalance and extra classes in the test set. 
Most of the software (like Weka) just ignore extra classes present in the test set. Doing this can skew the results especially when there is a huge class imbalance in the train set. So for multi-class classification with high-class imbalance and more classes in the test set than train set, the \textit{recall} will inherently be low. Therefore, while selecting the hyper-parameters for \textit{recall} averaging of the class labels should be avoided. This can be done by selecting 'macro' as a label average method for \textit{recall} in sklearn. This method calculates metrics for each label and finds their unweighted mean. Whereas \textit{precision} is not sensitive to the class labels; therefore, 'weighted' average method can be used (which finds the number of weighted instances for each label). 

\begin{figure*}[t]
  \centering
    \includegraphics[width=.8\textwidth]{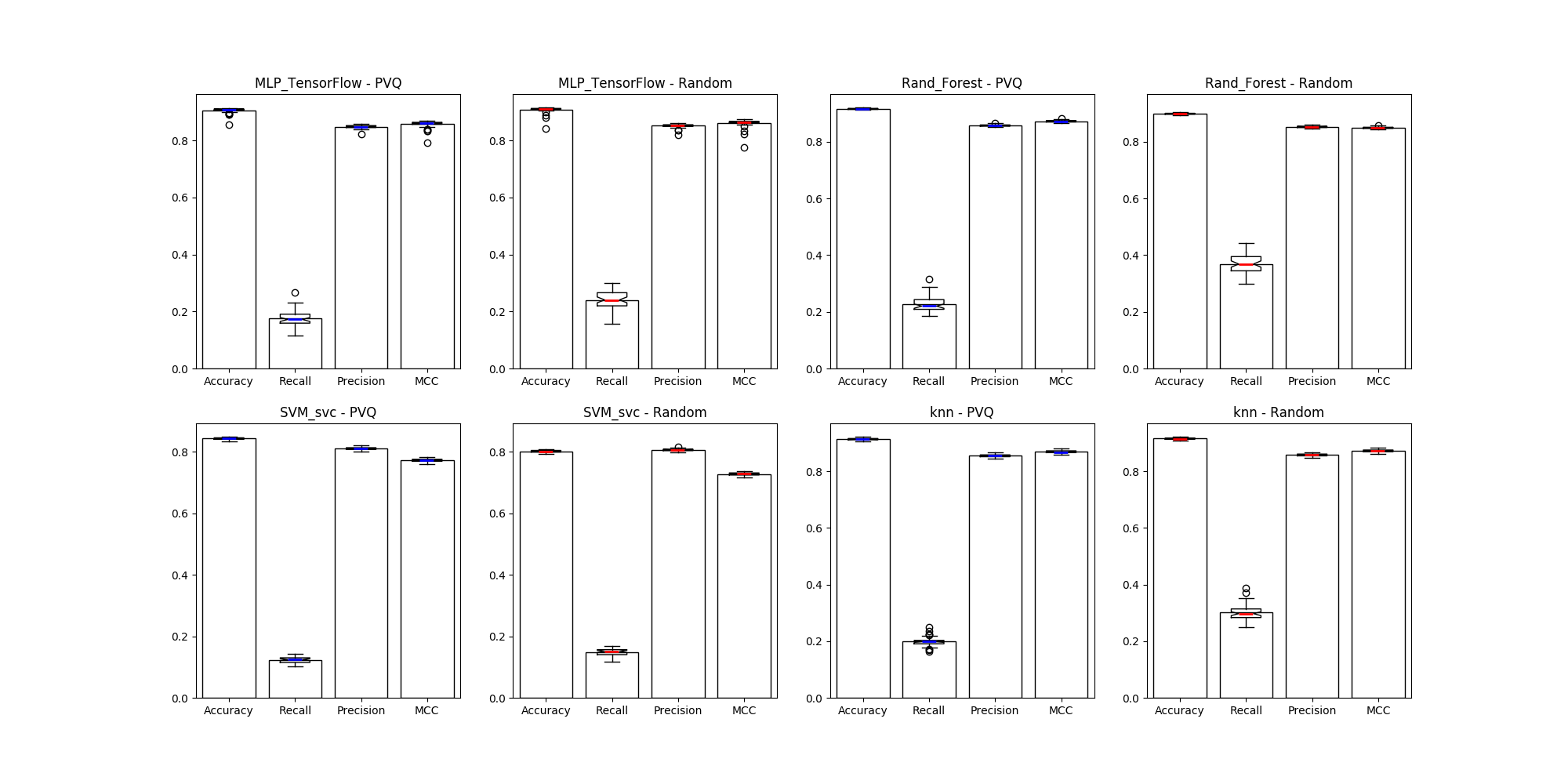}\\
\caption{\label{fig:box_plots_TF} The plot of Accuracy, Recall, Precision, and Mathews Correlation Coefficient of \textquotedblleft PVQ Sampled'' dataset (80k) vs. \textquotedblleft Randomly Sampled'' (80k), tested on 10k, randomly read from test set for different classifiers - 50 random runs. For comparison between randomly sampled and PVQ sampled, we appended "Random" and "PVQ" with each classifier. Top left pair is for MLP, top right pair is for RF, bottom left pair is for SVM, and bottom-right pair is for kNN.}
 \end{figure*}

For this group of experiments, we used two dense layered neural network architecture using Keras, to avoid overfitting. We used ADAM~\cite{kingma2014adam} learning rate optimizer. In the fist dense layer, we use \textit{Tanh} activation function and in the output layer we used \textit(sigmoid) activation function as this is a multi-class classification problem. The loss function used is categorical cross-entropy. We train these classifiers on PVQ sampled dataset and randomly selected dataset (about 80k) for 50 different iterations. Both models are tested on 10k test set randomly sampled from the full test set (which is 311k). The results show that PVQ sampling produces better accuracy, precision, and mcc for all classifiers. The results are shown in the box plot in Figure~\ref{fig:box_plots_TF}. 

\begin{figure}[!t]
\centering
   \includegraphics[width = .6\columnwidth]{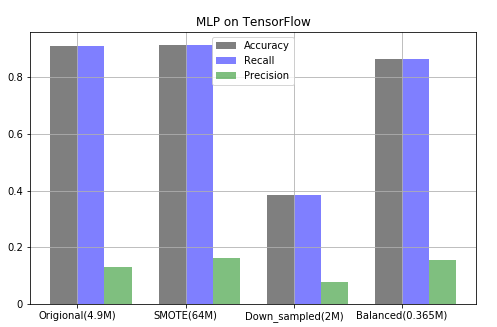}

\caption{\label{fig:Box_plots_comparison} Accuracy, Recall and Precision of: (a) Full KDDCUP dataset (4.9M) (b) SMOTE up-sampled dataset (64M) (c) Down-sampled dataset with "not-majority" class (2M) and (d) Up-sampled and down-Sampled to class "satan" (365k). Baseline created using MLP on TensorFlow backend for all 23 classes. Tested on full test set (311k) for 5 random runs.}
\end{figure}

% The run times for auto-sklearn are given in Figure~\ref{fig:runtimes} (right two columns), we can see auto-sklearn is a very expensive process and performs better for a small number of classes. Since the overall performance is still very low, we trained the classifiers manually as described in the next section.

% \subsubsection{Classification models built using vanilla hyper-parameters of selected classifiers (MLP with TensorFlow backend, RF, SVM and kNN)}\label{Manual}

% We use vanilla parameters of RF, SVM, and kNN classifiers. For MLP (on TensorFlow backend), we use two dense layered architecture using Keras, to avoid overfitting. We used ADAM~\cite{kingma2014adam} learning rate optimizer. In the fist dense layer, we use \textit{Tanh} activation function and in the output layer we use \textit(sigmoid) activation function as this is a multi-class classification problem. The loss function used is categorical cross-entropy. We train these classifiers first on PVQ sampled dataset (about 80k). Then, we randomly sample (80k) from the full dataset and built classification models for 50 different iterations. Both models are tested on 10k test set randomly sampled from the full test set (which is 311k). The results, which prove PVQ sampling is far superior than all other methods used, are shown in box plot in Figure~\ref{fig:box_plots_TF}, and discussion on the results is given in Section~\ref{discussion}.

%\begin{figure*}[ht]
%\centering
%    \includegraphics[height=8cm]{pvq_data2.png}
%    \caption{\label{fig:pvq_data2}KDDCUP Dataset}
%\end{figure*}

\section{Analysis and Discussion on Results} \label{discussion}

In this study, we analyzed cloud bound data streams with high-class imbalance. We processed the longer streams (one week of network traffic) as well as smaller streaming windows. We processed the data and built classification models to see how the class imbalance plays a role. We showed that the general approaches of classification model building from data streams either do not scale up or do not produce good results. Weka could not run SVM or any other complex classification other than Naive Bayes. We showed that MLP with TensorFlow backend (which uses parallel processing and chop the data into smaller subsets) does not perform well on an imbalanced dataset. Figure~\ref{fig:Box_plots_comparison} shows that, though we can use MLP on TensorFlow to analyze the \textit{full batch} in a reasonable time, the \textit{precision} is very low, which is not desirable result in this multi-class classification problem. 

This is because TensorFlow chops the data into smaller chunks in the background to build models in parallel. When the data are chopped, not all the classes can be selected/represented by any node in TensorFlow. This finding warrants a separate investigation as TensorFlow is currently used in many ML applications in the industry. Note that MLP on TensorFlow could only find a maximum of 16 classes, whereas PVQ was able to find 22 classes out of 23 classes present in the train set \ref{fig:random_classes}. Thus, using TensorFlow with classifiers does not help in building classification models from data streams even if the data stream window is fairly large (seven days or \textit{full batch} of KDDCUP in this case).

We also showed that the automated ML approaches like auto-sklearn either do not work on moderately sized datasets or take days to build classification models on reduced sized datasets. The reason is, the hyper-parameter optimization process (as explained in Section~\ref{results_group2a}) is very iterative and makes multiple copies of the dataset making, even moderate-sized dataset unsolvable. We found that auto-sklearn is a costly optimization process. With 80k random sampling for 50 iterations, it took over a week to suggest a model on the machine described in \ref{problem-statement}.

Figure \ref{fig:box_plots_TF} shows that \textquotedblleft PVQ sampled'' dataset (plots labeled with PVQ suffix) performed better in terms of model quality (higher \textit{accuracy}, \textit{precision}, and MCC) as compared to \textquotedblleft Randomly sampled'' dataset for all 50 runs for classifiers (RF, SVM, kNN, including MLP on TensorFlow). Low \textit{recall} for PVQ sampled data was also expected for a reason explained above. 

PVQ sampling was able to detect 22 out of 23 total classes in the train set (irrespective of the sampling window size), whereas a maximum of 16 classes distinguished by \textit{full batch} processing using TensorFlow. A spread of classes captured by each classification method when random sampling is used (irrespective of batch size) is shown in box plot Figure \ref{fig:random_classes}. 

In this publication, we described the PVQ sampling method, which improves the accuracy and performance of the classification algorithms for cloud-centric streaming data. We showed that PVQ removes class imbalance in the streaming data even with a small window. Most importantly, we showed that when graph-based parallel processing backends like TensorFlow are used with classification algorithms, the classifiers are not able to capture full classes present in a streaming dataset, even if a very large sampling window is used. Whereas, when data is PVQ sampled, the classification algorithms (with TensorFlow backend) capture almost all classes present in a dataset. 

\newcommand{\bibfont}{\footnotesize}

\bibliographystyle{IEEEtran}

\bibliography{refrences.bib}

\begin{thebibliography}{10}
\providecommand{\url}[1]{#1}
\csname url@samestyle\endcsname
\providecommand{\newblock}{\relax}
\providecommand{\bibinfo}[2]{#2}
\providecommand{\BIBentrySTDinterwordspacing}{\spaceskip=0pt\relax}
\providecommand{\BIBentryALTinterwordstretchfactor}{4}
\providecommand{\BIBentryALTinterwordspacing}{\spaceskip=\fontdimen2\font plus
\BIBentryALTinterwordstretchfactor\fontdimen3\font minus
  \fontdimen4\font\relax}
\providecommand{\BIBforeignlanguage}[2]{{%
\expandafter\ifx\csname l@#1\endcsname\relax
\typeout{** WARNING: IEEEtran.bst: No hyphenation pattern has been}%
\typeout{** loaded for the language `#1'. Using the pattern for}%
\typeout{** the default language instead.}%
\else
\language=\csname l@#1\endcsname
\fi
#2}}
\providecommand{\BIBdecl}{\relax}
\BIBdecl

\bibitem{lunardon2014rose}
N.~Lunardon, G.~Menardi, and N.~Torelli, ``Rose: A package for binary
  imbalanced learning.'' \emph{R Journal}, vol.~6, no.~1, 2014.

\bibitem{chawla2002smote}
N.~V. Chawla, K.~W. Bowyer, L.~O. Hall, and W.~P. Kegelmeyer, ``Smote:
  synthetic minority over-sampling technique,'' \emph{Journal of artificial
  intelligence research}, vol.~16, pp. 321--357, 2002.

\bibitem{chawla2003smoteboost}
N.~V. Chawla, A.~Lazarevic, L.~O. Hall, and K.~W. Bowyer, ``Smoteboost:
  Improving prediction of the minority class in boosting,'' in \emph{European
  conference on principles of data mining and knowledge discovery}.\hskip 1em
  plus 0.5em minus 0.4em\relax Springer, 2003, pp. 107--119.

\bibitem{kearns1998efficient}
M.~Kearns, ``Efficient noise-tolerant learning from statistical queries,''
  \emph{Journal of the ACM (JACM)}, vol.~45, no.~6, pp. 983--1006, 1998.

\bibitem{cup1999dataset}
K.~Cup, ``Dataset,'' \emph{available at the following website http://kdd. ics.
  uci. edu/databases/kddcup99/kddcup99. html}, vol.~72, 1999.

\bibitem{phua2010comprehensive}
C.~Phua, V.~Lee, K.~Smith, and R.~Gayler, ``A comprehensive survey of data
  mining-based fraud detection research,'' \emph{arXiv preprint
  arXiv:1009.6119}, 2010.

\bibitem{ring2017flow}
M.~Ring, S.~Wunderlich, D.~Gr{\"u}dl, D.~Landes, and A.~Hotho, ``Flow-based
  benchmark data sets for intrusion detection,'' in \emph{Proceedings of the
  16th European Conference on Cyber Warfare and Security. ACPI}, 2017, pp.
  361--369.

\bibitem{sharafaldin2018toward}
I.~Sharafaldin, A.~H. Lashkari, and A.~A. Ghorbani, ``Toward generating a new
  intrusion detection dataset and intrusion traffic characterization.'' in
  \emph{ICISSP}, 2018, pp. 108--116.

\bibitem{john1995estimating}
G.~H. John and P.~Langley, ``Estimating continuous distributions in bayesian
  classifiers,'' in \emph{Proceedings of the Eleventh conference on Uncertainty
  in artificial intelligence}.\hskip 1em plus 0.5em minus 0.4em\relax Morgan
  Kaufmann Publishers Inc., 1995, pp. 338--345.

\bibitem{NIPS2015_5872}
\BIBentryALTinterwordspacing
M.~Feurer, A.~Klein, K.~Eggensperger, J.~Springenberg, M.~Blum, and F.~Hutter,
  ``Efficient and robust automated machine learning,'' in \emph{Advances in
  Neural Information Processing Systems 28}, C.~Cortes, N.~D. Lawrence, D.~D.
  Lee, M.~Sugiyama, and R.~Garnett, Eds.\hskip 1em plus 0.5em minus 0.4em\relax
  Curran Associates, Inc., 2015, pp. 2962--2970. [Online]. Available:
  \url{http://papers.nips.cc/paper/5872-efficient-and-robust-automated-machine-learning.pdf}
\BIBentrySTDinterwordspacing

\bibitem{zoller2019survey}
M.-A. Z{\"o}ller and M.~F. Huber, ``Survey on automated machine learning,''
  \emph{arXiv preprint arXiv:1904.12054}, 2019.

\bibitem{drucker1994boosting}
H.~Drucker, C.~Cortes, L.~D. Jackel, Y.~LeCun, and V.~Vapnik, ``Boosting and
  other ensemble methods,'' \emph{Neural Computation}, vol.~6, no.~6, pp.
  1289--1301, 1994.

\bibitem{he2010parallel}
Q.~He, F.~Zhuang, J.~Li, and Z.~Shi, ``Parallel implementation of
  classification algorithms based on mapreduce,'' in \emph{International
  Conference on Rough Sets and Knowledge Technology}.\hskip 1em plus 0.5em
  minus 0.4em\relax Springer, 2010, pp. 655--662.

\bibitem{chen2017parallel}
J.~Chen, K.~Li, Z.~Tang, K.~Bilal, S.~Yu, C.~Weng, and K.~Li, ``A parallel
  random forest algorithm for big data in a spark cloud computing
  environment,'' \emph{IEEE Transactions on Parallel \& Distributed Systems},
  no.~1, pp. 1--1, 2017.

\bibitem{abadi2016tensorflow}
M.~Abadi, P.~Barham, J.~Chen, Z.~Chen, A.~Davis, J.~Dean, M.~Devin,
  S.~Ghemawat, G.~Irving, M.~Isard \emph{et~al.}, ``Tensorflow: a system for
  large-scale machine learning.'' in \emph{OSDI}, vol.~16, 2016, pp. 265--283.

\bibitem{seide2016cntk}
F.~Seide and A.~Agarwal, ``Cntk: Microsoft's open-source deep-learning
  toolkit,'' in \emph{Proceedings of the 22nd ACM SIGKDD International
  Conference on Knowledge Discovery and Data Mining}.\hskip 1em plus 0.5em
  minus 0.4em\relax ACM, 2016, pp. 2135--2135.

\bibitem{bastien2012theano}
F.~Bastien, P.~Lamblin, R.~Pascanu, J.~Bergstra, I.~Goodfellow, A.~Bergeron,
  N.~Bouchard, D.~Warde-Farley, and Y.~Bengio, ``Theano: new features and speed
  improvements,'' \emph{arXiv preprint arXiv:1211.5590}, 2012.

\bibitem{akidau2015dataflow}
T.~Akidau, R.~Bradshaw, C.~Chambers, S.~Chernyak, R.~J.
  Fern{\'a}ndez-Moctezuma, R.~Lax, S.~McVeety, D.~Mills, F.~Perry, E.~Schmidt
  \emph{et~al.}, ``The dataflow model: a practical approach to balancing
  correctness, latency, and cost in massive-scale, unbounded, out-of-order data
  processing,'' \emph{Proceedings of the VLDB Endowment}, vol.~8, no.~12, pp.
  1792--1803, 2015.

\bibitem{zeek}
\BIBentryALTinterwordspacing
``Zeek.'' [Online]. Available: \url{https://www.zeek.org/}
\BIBentrySTDinterwordspacing

\bibitem{snort}
\BIBentryALTinterwordspacing
``Snort.'' [Online]. Available: \url{https://www.snort.org/}
\BIBentrySTDinterwordspacing

\bibitem{anton2018evaluation}
S.~D. Anton, S.~Kanoor, D.~Fraunholz, and H.~D. Schotten, ``Evaluation of
  machine learning-based anomaly detection algorithms on an industrial
  modbus/tcp data set,'' in \emph{Proceedings of the 13th International
  Conference on Availability, Reliability and Security}.\hskip 1em plus 0.5em
  minus 0.4em\relax ACM, 2018, p.~41.

\bibitem{gama2010knowledge}
J.~Gama, \emph{Knowledge discovery from data streams}.\hskip 1em plus 0.5em
  minus 0.4em\relax Chapman and Hall/CRC, 2010.

\bibitem{gaber2005mining}
M.~M. Gaber, A.~Zaslavsky, and S.~Krishnaswamy, ``Mining data streams: a
  review,'' \emph{ACM Sigmod Record}, vol.~34, no.~2, pp. 18--26, 2005.

\bibitem{pratama2014pclass}
M.~Pratama, S.~G. Anavatti, M.~Joo, and E.~D. Lughofer, ``pclass: an effective
  classifier for streaming examples,'' \emph{IEEE Transactions on Fuzzy
  Systems}, vol.~23, no.~2, pp. 369--386, 2014.

\bibitem{silva2013data}
J.~A. Silva, E.~R. Faria, R.~C. Barros, E.~R. Hruschka, A.~C. De~Carvalho, and
  J.~Gama, ``Data stream clustering: A survey,'' \emph{ACM Computing Surveys
  (CSUR)}, vol.~46, no.~1, p.~13, 2013.

\bibitem{leevy2018survey}
J.~L. Leevy, T.~M. Khoshgoftaar, R.~A. Bauder, and N.~Seliya, ``A survey on
  addressing high-class imbalance in big data,'' \emph{Journal of Big Data},
  vol.~5, no.~1, p.~42, 2018.

\bibitem{mehta2017concept}
S.~Mehta \emph{et~al.}, ``Concept drift in streaming data classification:
  Algorithms, platforms and issues,'' \emph{Procedia computer science}, vol.
  122, pp. 804--811, 2017.

\bibitem{pratama2016incremental}
M.~Pratama, J.~Lu, E.~Lughofer, G.~Zhang, and M.~J. Er, ``An incremental
  learning of concept drifts using evolving type-2 recurrent fuzzy neural
  networks,'' \emph{IEEE Transactions on Fuzzy Systems}, vol.~25, no.~5, pp.
  1175--1192, 2016.

\bibitem{sayed2012learning}
M.~Sayed-Mouchaweh and E.~Lughofer, \emph{Learning in non-stationary
  environments: methods and applications}.\hskip 1em plus 0.5em minus
  0.4em\relax Springer Science \& Business Media, 2012.

\bibitem{gama2014survey}
J.~Gama, I.~{\v{Z}}liobait{\.e}, A.~Bifet, M.~Pechenizkiy, and A.~Bouchachia,
  ``A survey on concept drift adaptation,'' \emph{ACM computing surveys
  (CSUR)}, vol.~46, no.~4, p.~44, 2014.

\bibitem{bifet2015streamdm}
A.~Bifet, S.~Maniu, J.~Qian, G.~Tian, C.~He, and W.~Fan, ``Streamdm: Advanced
  data mining in spark streaming,'' in \emph{2015 IEEE International Conference
  on Data Mining Workshop (ICDMW)}.\hskip 1em plus 0.5em minus 0.4em\relax
  IEEE, 2015, pp. 1608--1611.

\bibitem{malhotra2015long}
P.~Malhotra, L.~Vig, G.~Shroff, and P.~Agarwal, ``Long short term memory
  networks for anomaly detection in time series,'' in \emph{Proceedings}.\hskip
  1em plus 0.5em minus 0.4em\relax Presses universitaires de Louvain, 2015,
  p.~89.

\bibitem{johnson2019survey}
J.~M. Johnson and T.~M. Khoshgoftaar, ``Survey on deep learning with class
  imbalance,'' \emph{Journal of Big Data}, vol.~6, no.~1, p.~27, 2019.

\bibitem{akouemo2016probabilistic}
H.~N. Akouemo and R.~J. Povinelli, ``Probabilistic anomaly detection in natural
  gas time series data,'' \emph{International Journal of Forecasting}, vol.~32,
  no.~3, pp. 948--956, 2016.

\bibitem{al2015efficient}
O.~Y. Al-Jarrah, P.~D. Yoo, S.~Muhaidat, G.~K. Karagiannidis, and K.~Taha,
  ``Efficient machine learning for big data: A review,'' \emph{Big Data
  Research}, vol.~2, no.~3, pp. 87--93, 2015.

\bibitem{xing2015petuum}
E.~P. Xing, Q.~Ho, W.~Dai, J.~K. Kim, J.~Wei, S.~Lee, X.~Zheng, P.~Xie,
  A.~Kumar, and Y.~Yu, ``Petuum: A new platform for distributed machine
  learning on big data,'' \emph{IEEE Transactions on Big Data}, vol.~1, no.~2,
  pp. 49--67, 2015.

\bibitem{he2008learning}
H.~He and E.~A. Garcia, ``Learning from imbalanced data,'' \emph{IEEE
  Transactions on Knowledge \& Data Engineering}, no.~9, pp. 1263--1284, 2008.

\bibitem{apachespark}
\BIBentryALTinterwordspacing
``Apache spark project.'' [Online]. Available: \url{http://spark.apache.org/}
\BIBentrySTDinterwordspacing

\bibitem{apacheflink}
\BIBentryALTinterwordspacing
``Apache flink project.'' [Online]. Available: \url{http://flink.apache.org/}
\BIBentrySTDinterwordspacing

\bibitem{chintapalli2016benchmarking}
S.~Chintapalli, D.~Dagit, B.~Evans, R.~Farivar, T.~Graves, M.~Holderbaugh,
  Z.~Liu, K.~Nusbaum, K.~Patil, B.~J. Peng \emph{et~al.}, ``Benchmarking
  streaming computation engines: Storm, flink and spark streaming,'' in
  \emph{2016 IEEE international parallel and distributed processing symposium
  workshops (IPDPSW)}.\hskip 1em plus 0.5em minus 0.4em\relax IEEE, 2016, pp.
  1789--1792.

\bibitem{gersho2012vector}
A.~Gersho and R.~M. Gray, \emph{Vector quantization and signal
  compression}.\hskip 1em plus 0.5em minus 0.4em\relax Springer Science \&
  Business Media, 2012, vol. 159.

\bibitem{gray1998quantization}
R.~M. Gray and D.~L. Neuhoff, ``Quantization,'' \emph{IEEE transactions on
  information theory}, vol.~44, no.~6, pp. 2325--2383, 1998.

\bibitem{dirichlet1850reduction}
G.~L. Dirichlet, ``{\"U}ber die reduction der positiven quadratischen formen
  mit drei unbestimmten ganzen zahlen.'' \emph{Journal f{\"u}r die reine und
  angewandte Mathematik}, vol.~40, pp. 209--227, 1850.

\bibitem{voronoi1908nouvelles}
G.~Vorono{\"\i}, ``Nouvelles applications des param{\`e}tres continus {\`a} la
  th{\'e}orie des formes quadratiques. deuxi{\`e}me m{\'e}moire. recherches sur
  les parall{\'e}llo{\`e}dres primitifs.'' \emph{Journal f{\"u}r die reine und
  angewandte Mathematik}, vol. 134, pp. 198--287, 1908.

\bibitem{lloyd1982least}
S.~Lloyd, ``Least squares quantization in pcm,'' \emph{IEEE transactions on
  information theory}, vol.~28, no.~2, pp. 129--137, 1982.

\bibitem{forgy1965cluster}
E.~W. Forgy, ``Cluster analysis of multivariate data: efficiency versus
  interpretability of classifications,'' \emph{biometrics}, vol.~21, pp.
  768--769, 1965.

\bibitem{nakamura2013lvq}
M.~Nakamura, Y.~Kajiwara, A.~Otsuka, and H.~Kimura, ``Lvq-smote--learning
  vector quantization based synthetic minority over--sampling technique for
  biomedical data,'' \emph{BioData mining}, vol.~6, no.~1, p.~16, 2013.

\bibitem{khoshrou2015learning}
S.~Khoshrou, J.~S. Cardoso, and L.~F. Teixeira, ``Learning from evolving video
  streams in a multi-camera scenario,'' \emph{Machine Learning}, vol. 100, no.
  2-3, pp. 609--633, 2015.

\bibitem{vesanto1999self}
J.~Vesanto, J.~Himberg, E.~Alhoniemi, J.~Parhankangas \emph{et~al.},
  ``Self-organizing map in matlab: the som toolbox,'' in \emph{Proceedings of
  the Matlab DSP conference}, vol.~99, 1999, pp. 16--17.

\bibitem{kohonen2013essentials}
T.~Kohonen, ``Essentials of the self-organizing map,'' \emph{Neural networks},
  vol.~37, pp. 52--65, 2013.

\bibitem{vesanto2000neural}
J.~Vesanto, ``Neural network tool for data mining: Som toolbox,'' in
  \emph{Proceedings of symposium on tool environments and development methods
  for intelligent systems (TOOLMET2000)}.\hskip 1em plus 0.5em minus
  0.4em\relax Citeseer, 2000, pp. 184--196.

\bibitem{bishop2012pattern}
C.~M. Bishop, \emph{Pattern recognition and machine learning}.\hskip 1em plus
  0.5em minus 0.4em\relax Springer Science, 2006.

\bibitem{aha1991instance}
D.~W. Aha, D.~Kibler, and M.~K. Albert, ``Instance-based learning algorithms,''
  \emph{Machine learning}, vol.~6, no.~1, pp. 37--66, 1991.

\bibitem{breiman2001random}
L.~Breiman, ``Random forests,'' \emph{Machine learning}, vol.~45, no.~1, pp.
  5--32, 2001.

\bibitem{rumelhart1985learning}
D.~E. Rumelhart, G.~E. Hinton, and R.~J. Williams, ``Learning internal
  representations by error propagation,'' California Univ San Diego La Jolla
  Inst for Cognitive Science, Tech. Rep., 1985.

\bibitem{lippmann2000evaluating}
R.~P. Lippmann, D.~J. Fried, I.~Graf, J.~W. Haines, K.~R. Kendall, D.~McClung,
  D.~Weber, S.~E. Webster, D.~Wyschogrod, R.~K. Cunningham \emph{et~al.},
  ``Evaluating intrusion detection systems: The 1998 darpa off-line intrusion
  detection evaluation,'' in \emph{DARPA Information Survivability Conference
  and Exposition, 2000. DISCEX'00. Proceedings}, vol.~2.\hskip 1em plus 0.5em
  minus 0.4em\relax IEEE, 2000, pp. 12--26.

\bibitem{stolfo2000cost}
S.~J. Stolfo, W.~Fan, W.~Lee, A.~Prodromidis, and P.~K. Chan, ``Cost-based
  modeling for fraud and intrusion detection: Results from the jam project,''
  COLUMBIA UNIV NEW YORK DEPT OF COMPUTER SCIENCE, Tech. Rep., 2000.

\bibitem{bishop2006pattern}
C.~M. Bishop, \emph{Pattern recognition and machine learning}.\hskip 1em plus
  0.5em minus 0.4em\relax springer, 2006.

\bibitem{mackay2003information}
D.~J. MacKay and D.~J. Mac~Kay, \emph{Information theory, inference and
  learning algorithms}.\hskip 1em plus 0.5em minus 0.4em\relax Cambridge
  university press, 2003.

\bibitem{sahu2014detail}
S.~K. Sahu, S.~Sarangi, and S.~K. Jena, ``A detail analysis on intrusion
  detection datasets,'' in \emph{Advance Computing Conference (IACC), 2014 IEEE
  International}.\hskip 1em plus 0.5em minus 0.4em\relax IEEE, 2014, pp.
  1348--1353.

\bibitem{kingma2014adam}
D.~P. Kingma and J.~Ba, ``Adam: A method for stochastic optimization,''
  \emph{arXiv preprint arXiv:1412.6980}, 2014.

\bibitem{chollet2015keras}
F.~Chollet \emph{et~al.}, ``Keras,'' \url{https://keras.io}, 2015.

\bibitem{gorodkin2004comparing}
J.~Gorodkin, ``Comparing two k-category assignments by a k-category correlation
  coefficient,'' \emph{Computational biology and chemistry}, vol.~28, no. 5-6,
  pp. 367--374, 2004.

\bibitem{powers2011evaluation}
D.~POWERS, ``Evaluation: From precision, recall and f-measure to roc.,
  informedness, markedness \& correlation,'' \emph{JOURNAL OF MACHINE LEARNING
  TECHNOLOGIES}, 2011.

\bibitem{boughorbel2017optimal}
S.~Boughorbel, F.~Jarray, and M.~El-Anbari, ``Optimal classifier for imbalanced
  data using matthews correlation coefficient metric,'' \emph{PloS one},
  vol.~12, no.~6, p. e0177678, 2017.

\bibitem{tavallaee2009detailed}
M.~Tavallaee, E.~Bagheri, W.~Lu, and A.~A. Ghorbani, ``A detailed analysis of
  the kdd cup 99 data set,'' in \emph{2009 IEEE Symposium on Computational
  Intelligence for Security and Defense Applications}.\hskip 1em plus 0.5em
  minus 0.4em\relax IEEE, 2009, pp. 1--6.

\bibitem{smac}
\BIBentryALTinterwordspacing
``Smac: Sequential model-based algorithm configuration.'' [Online]. Available:
  \url{http://www.cs.ubc.ca/labs/beta/Projects/SMAC/}
\BIBentrySTDinterwordspacing

\bibitem{pedregosa2011scikit}
F.~Pedregosa, G.~Varoquaux, A.~Gramfort, V.~Michel, B.~Thirion, O.~Grisel,
  M.~Blondel, P.~Prettenhofer, R.~Weiss, V.~Dubourg \emph{et~al.},
  ``Scikit-learn: Machine learning in python,'' \emph{Journal of machine
  learning research}, vol.~12, no. Oct, pp. 2825--2830, 2011.

\end{thebibliography}
%\bibliography{refrences.bib}

\end{document}